%% file: lncs_format.tex
\documentclass[runningheads]{llncs}
\usepackage{graphicx}
\input{settings}%

\makeatletter
\usepackage{xspace}
\DeclareRobustCommand\onedot{\futurelet\@let@token\@onedot}
\def\@onedot{\ifx\@let@token.\else.\null\fi\xspace}
\def\ie{\emph{i.e}\onedot} 
\def\etal{\emph{et al}\onedot}
\makeatother

\begin{document}
\title{Can masking background and object reduce static bias for zero-shot action recognition?}
\titlerunning{Masking background and object reduce static bias}

\author{
Takumi Fukuzawa\inst{1}\orcidID{0009-0008-8581-8261}
\and
Kensho Hara\inst{2}\orcidID{0000-0001-6463-7738}
\and
Hirokatsu Kataoka\inst{2}\orcidID{0000-0001-8844-165X}
\and \\
Toru Tamaki\inst{1}\orcidID{0000-0001-9712-7777}
}

\authorrunning{T. Fukuzawa et al.}

\institute{
Nagoya Institute of Technology, Japan\\
\email{t.fukuzawa.986@nitech.jp},
\email{tamaki.toru@nitech.ac.jp}
\and
National Institute of Advanced Industrial Science and Technology (AIST), Japan\\
\email{hirokatsu.kataoka@aist.go.jp},
\email{kensho.hara@aist.go.jp}
}

\maketitle              

\begin{abstract}
\input{abstract}

\end{abstract}

\input{main_text.tex}

\section*{Acknowledgements}
This work was supported in part by JSPS KAKENHI Grant Number JP22K12090.

\bibliographystyle{splncs04}
\bibliography{all,mybib}

\end{document}

%% file: settings.tex
\usepackage{graphicx}
\usepackage[sort,space,adjust,nocompress]{cite}

\usepackage[breaklinks=true,colorlinks,bookmarks=false]{hyperref}
\usepackage{amsmath}
\usepackage{amssymb}

\usepackage{multirow}
\usepackage{multicol}
\usepackage{diagbox}

\usepackage[labelformat=simple]{subcaption}

\usepackage{algorithmic}
\usepackage{algorithm}

\usepackage{caption}
\newcommand{\ve}{\boldsymbol{v}}
\newcommand{\te}{\boldsymbol{t}}


%% file: abstract.tex
In this paper, we address the issue of static bias in zero-shot action recognition. Action recognition models need to represent the action itself, not the appearance. However, some fully-supervised works show that models often rely on static appearances, such as the background and objects, rather than human actions. This issue, known as static bias, has not been investigated for zero-shot. Although CLIP-based zero-shot models are now common, it remains unclear if they sufficiently focus on human actions, as CLIP primarily captures appearance features related to languages. In this paper, we investigate the influence of static bias in zero-shot action recognition with CLIP-based models.
Our approach involves masking backgrounds, objects, and people differently during training and validation. Experiments with masking background show that models depend on background bias as their performance decreases for Kinetics400. However, for Mimetics, which has a weak background bias, masking the background leads to improved performance even if the background is masked during validation. Furthermore, masking both the background and objects in different colors improves performance for SSv2, which has a strong object bias. These results suggest that masking the background or objects during training prevents models from overly depending on static bias and makes them focus more on human action.

%% file: main_text.tex
\section{Introduction}

Action recognition is a task for recognizing the actions of a person in a video and is expected to have various applications \cite{Hara_IEICE_ED2020_Action_Recognition_Survey,Kong_IJCV2022_Action_Recognition_Survey,Ulhaq_arXiv2022_Transformers_Action_Recognition_Survey}. Supervised learning of action recognition models requires a large dataset \cite{kay_arXiv2017_kinetics400,Soomro_arXiv2012_UCF101,Kuehne_ICCV2011_HMDB51}, however, collecting such a dataset can be costly, and the amount of collectable training data is often limited in real-world applications. For this reason, zero-shot action recognition models have been proposed \cite{Estevam_Neurocomputing2021_zero-shot_action_recog_survey}.
Zero-shot learning involves predicting unseen categories during testing, and a current popular approach is to 
use VLM models such as CLIP \cite{Radford_2021_CLIP} for
semantic proximity between visual elements and category name texts in latent space \cite{Ma_2022_XCLIP,MAXI_Lin_2023_ICCV,BIKE_Wu_2023_CVPR,Rasheed_2023_ViFiCLIP}.

Action recognition requires a representation of the action itself. However, as reported by \cite{Action_swap}, many existing methods tend to rely on static appearances, such as the background of the scene and objects in the scene, rather than human actions. This issue is called \emph{static bias}, often referred to as \emph{background bias} or \emph{object bias} in action recognition tasks, commonly known as representation bias \cite{RESOUND_Li_2018_ECCV,Hara_2021_CVPRW}.
For practical applications, an action recognition system should be able to correctly predict actions even with a strong background bias; for example, dancing on a basketball court should be classified as ``dancing'', not ``playing basketball''.
This issue can cause incorrect predictions, especially in zero-shot learning where training and validation categories differ, and is more critical than in fully-supervised action recognition.

Datasets commonly used for action recognition, such as Kinetics400 \cite{kay_arXiv2017_kinetics400} and UCF101 \cite{Soomro_arXiv2012_UCF101}, are known to have a strong \emph{background bias}. For example, many videos in the category ``Hitting a baseball'' include substantial background information such as the baseball ground.
A single frame of videos often contains sufficient information to predict the action category \cite{Zhu_arXiv2020_Action_Recognition_Survey}, which leads to models becoming overly dependent on background bias.
Chung \etal \cite{Action_swap} demonstrated that experiments involving training on ``background-only'' videos of Kinetics400, with the person inpainted out, showed that many action recognition models perform about 50\% in accuracy.
Even when the person in the scene is masked with a single-colored rectangle, models are reported to still be able to predict categories reasonably \cite{without_human,Weinzaepfel_IJCV2021_Mimetics_dataset,Choi_NEURIPS2019_Dance_in_the_Mall}.
This background bias has been extensively analyzed in the context of fully-supervised action recognition \cite{without_human,Weinzaepfel_IJCV2021_Mimetics_dataset,Action_swap,luo-etal2022-DARK,Choi_NEURIPS2019_Dance_in_the_Mall}. However, it is still uncertain for zero-shot action recognition with VLM models that extend CLIP, which is proficient at representing appearance features related to languages.

In contrast, Something-Something v2 (SSv2) \cite{Goyal_2017ICCV_ssv2} exhibits a weak background bias.
The categories are represented as sentences with a placeholder [something]
and the background is so simple that there are hardly any clues.
As a result, it needs to focus on temporal changes such as human posture, hand shape, and object placement across multiple frames.
However, SSv2 has a significant \emph{object bias} and models tend to rely on objects that appear frequently. This bias is particularly evident in categories like ``Tearing [something] into two pieces''; tearable objects such as paper rarely appear in other categories, so having just paper in the scene leads to predicting that category.
This object bias might cause poorer performance for unseen categories \cite{MAXI_Lin_2023_ICCV, Rasheed_2023_ViFiCLIP}. 
Therefore, it is necessary to evaluate the behavior of zero-shot models using CLIP for object bias.

Previous \emph{fully-supervised} studies have addressed static bias by masking the background \cite{Action_swap} or objects \cite{luo-etal2022-DARK}. Some methods use a bounding box (bbox) to mask the person \cite{Choi_NEURIPS2019_Dance_in_the_Mall} or the object \cite{luo-etal2022-DARK}. Others mask and replace the background with another video background \cite{Action_swap, sugiura_VISAPP2024_s3aug}.
Inspired by these masking approaches, we use two masking approaches to investigate background and object biases for \emph{zero-shot} action recognition, as shown in Figure \ref{fig:masking approaches}.
Our approach for background bias masks the background with random colors to remove any potential clues. This makes the model focus solely on the person in action, similar to \cite{Action_swap}.
To address object bias, we mask objects using either the object's bounding box, its shape, or the entire background. In either case, human regions are excluded.
Masking objects with bounding boxes removes the object's shape information, whereas masking the object's shape eliminates its texture but retains its shape. Masking the whole background can effectively remove object bias; however, we opt to keep the object's shape, as completely removing it may not always be effective during training.

In the experiment, we introduce new evaluation metrics, B-top1 and P-top1, to assess whether the model's predictions are based on the background or person regions. The experimental results confirmed that using the masking approaches improved P-top1, a performance indicator focusing on person regions, against both background and object bias.

\section{Related work}

\subsection{Action recognition}

Action recognition involves predicting human actions,
and various methods have been proposed \cite{Hara_IEICE_ED2020_Action_Recognition_Survey,Kong_IJCV2022_Action_Recognition_Survey,Selva_arXiv2022_Video_Trans_Survey,Simonyan_NIPS2014_Two-Stream,Feichtenhofer_2020CVPR_X3D,Arnab_2021_ICCV_ViVit,Carreira_2017CVPR_I3D}.
Famous datasets used in this task include
HMDB51\cite{Kuehne_ICCV2011_HMDB51},
UCF101 \cite{Soomro_arXiv2012_UCF101}, and
Kinetics400 \cite{kay_arXiv2017_kinetics400}.
These datasets contain videos that last from a few seconds to several dozen seconds,
and their action categories often comprise a few words, which makes them suitable for zero-shot setting as well as fully-supervised settings.
SSv2 \cite{Goyal_2017ICCV_ssv2} is another commonly used dataset,
and its categories are given as template sentences.
The background is relatively simple, so it is necessary to observe the temporal changes in the video.

\subsection{Zero-shot action recognition}

Zero-shot action recognition has also been studied extensively,
often using CLIP \cite{Radford_2021_CLIP} with the ability to compute the similarity between images and texts in the embedding space.
X-CLIP \cite{Ma_2022_XCLIP} and ActionCLIP \cite{Wang_2021_ActionCLIP} 
introduced temporal interaction between frames by using interframe attention or temporal transformers, while ViFi-CLIP \cite{Rasheed_2023_ViFiCLIP} simply averages CLIP embeddings of each frame with prompt learning to preserve the pre-trained CLIP features.
These VLM-based models work well for zero-shot settings with datasets, such as Kinetics400, which have label texts that include names of objects that appear in the scene. However, the performance of SSv2 is not as good as that of Kinetics400
because the labels are given as templates without object names.
In addition, no zero-shot methods have explicitly considered background and object biases.

\subsection{Analyzing static bias}

There is prior work that aims to analyze and counteract background and object biases for fully-supervised action recognition.
Action-Swap \cite{Action_swap} removes background bias by extracting the person region from a video and pasting it onto different backgrounds from videos of different actions. It also proposes a metric to measure how much a model focuses on actions by evaluating performance with these background-swapped videos. S3Aug \cite{sugiura_VISAPP2024_s3aug} generates diverse backgrounds using a generative model while preserving the person region. Choi \etal \cite{Choi_NEURIPS2019_Dance_in_the_Mall} propose scene-independent action features by masking a person with bounding boxes and separating the background and person information. DARK \cite{luo-etal2022-DARK} masks objects with bounding boxes and predicts action verbs and nouns in separate branches.

In contrast, we address the open question for zero-shot action recognition: How are VLM models that utilize language affected by static bias? This has not been considered yet.

\section{Masking approaches}

In this section, we introduce two masking approaches: one for addressing background bias and the other for addressing object bias. In the following, masking
by a binary mask $M_i \in \{0, 1\}^{H \times W}$ is represented as:
\begin{align}
X_i = \overline{M}_i \odot V_i + \mathbf{c} \cdot M_i,
\quad i=1, \ldots, T,
\label{eq:mask}
\end{align}
where
$V_i \in R^{3 \times H \times W}$ is an RGB frame of a given video clip with $T$ frames of height $H$ and width $W$,
and $X_i$ is the masked frame.
The complement $\overline{M}_i = \mathbf{1} - M_i$ denotes the flipping of the mask values, where $\mathbf{1} \in \{1\}^{H \times W}$ is a mask filled with ones.
In the following, we omit the frame index $i$ because there is no confusion.
The 3d vector $\mathbf{c} \in R^3$ represents the masking color,
$\odot$ denotes an element-wise multiplication with dimension alignment between the mask and the frame, and
$\cdot$ represents a tensor product of the color vector and the mask to generate an RGB image.

If performance decreases when models are trained with the background or objects masked compared to when they are not masked,
it suggests that they depend on background or object biases \cite{RESOUND_Li_2018_ECCV,Hara_2021_CVPRW,Action_swap}.

\subsection{Masking for background bias}
\label{sec:Masking for background bias}

\begin{figure}[t]
    \centering

    \begin{minipage}[t]{0.30\linewidth}
        \centering
        \includegraphics[clip, width=\textwidth]{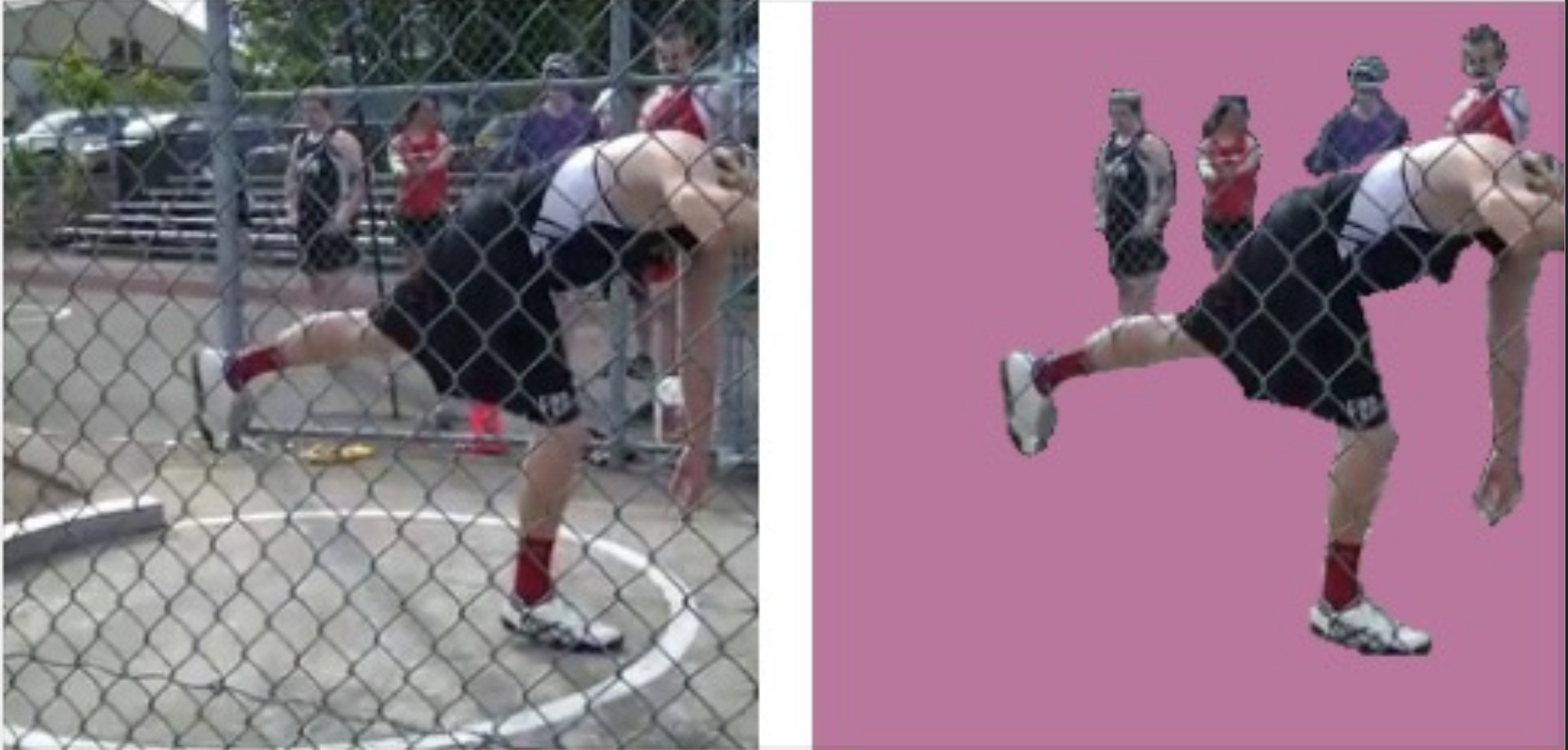}
        \subcaption{}
        \label{fig:mask_k400}
    \end{minipage}
    \hspace{0.05\linewidth}
    \begin{minipage}[t]{0.60\linewidth}
        \begin{center}
            \includegraphics[clip, width=\textwidth]{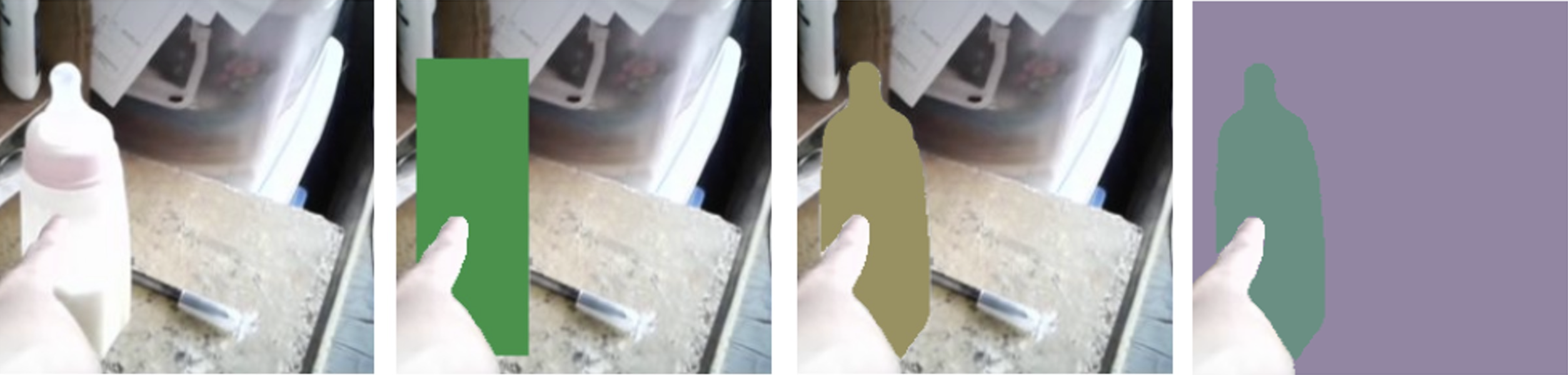}
            \subcaption{}
            \label{fig:mask_ssv2}
        \end{center}
    \end{minipage}

    \caption{Examples of masking.
    (a) Masking background.
    (left) Original frames, and
    (right) masking with sampled colors.
    (b) Masking objects.
    From left to right,
    original frames,
    masking the object with bbox and its shape, and
    masking the background and object.
    }
    \label{fig:masking approaches}

\end{figure}

To mitigate background bias, we mask the background of each frame to eliminate any potential clues from the background. This approach forces the model to focus only on the foreground, that is, the person performing the action. Instance segmentation is used to create a mask $M_\mathrm{person}^\mathrm{shape} \in \{0, 1\}^{H \times W}$, assigning 1 to pixels in the human region and 0 otherwise. The human region is preserved while the background is masked as follows:
\begin{align}
    M_\mathrm{bg} 
    &= \overline{M_\mathrm{person}^\mathrm{shape}},
    \quad
    X
    = \overline{M_\mathrm{bg}} \odot V 
    + \mathbf{c} \cdot M_\mathrm{bg}.
\label{eq:bg_mask}
\end{align}
This approach is similar to the background masking in \cite{Action_swap},
however, we do not use a fixed color for masking
because the same masking color for all videos might not be useful for learning features.
For each video, we sample an RGB color $\mathbf{c} \sim \mathcal{N}(\mathbf{0}, I)$ from a standard normal distribution $\mathcal{N}$ with mean $\mathbf{0}$ and unit standard deviation $I$, assuming that the pixel values are normalized.
This results in different videos having their backgrounds masked in different colors, as shown in Figure \ref{fig:mask_k400}.

\subsection{Masking for object bias}
\label{sec:Masking for object bias}

To eliminate object bias, we mask objects in three ways:
the object's bounding box, its shape, or the entire background (Figure \ref{fig:mask_ssv2}).

\subsubsection{Masking object bounding box.}

The first is ``object bounding box masking'' which masks objects with bounding boxes. This removes the shape information of the objects while also hiding a small part of the background.
Note that no masking is applied to human regions within the bounding box.

Let $M_\mathrm{object}^\mathrm{bbox} \in \{0, 1\}^{H \times W}$ be a mask of the bounding boxes of the object, assigning 1 to the pixels inside the bounding boxes and 0 otherwise.
Then the masking is done by;
\begin{align}
    M_\mathrm{object}^\mathrm{bbox'}
    &= \overline{M_\mathrm{object}^\mathrm{bbox}} \cup M_\mathrm{person}^\mathrm{shape},
    \quad
    X
    = M_\mathrm{object}^\mathrm{bbox'} \odot V
    + \mathbf{c} \cdot \overline{M_\mathrm{object}^\mathrm{bbox'}},
    \label{eq:mask_ob}
\end{align}
where $\cup$ is element-wise logical OR.

\subsubsection{Masking object shape.}

The second is ``object shape masking''. This masks the object regions, not the bounding box.
While it does not hide non-object areas, the shape of objects can still be visible through the mask's shape.

Let $M_\mathrm{object}^\mathrm{shape} \in \{0, 1\}^{H \times W}$ be a mask of the regions of the objects, assigning 1 to pixels inside the regions and 0 otherwise.
Then the masking is done by;
\begin{align}
    X
    &= \overline{M_\mathrm{object}^\mathrm{shape}} \odot V
    + \mathbf{c} \cdot M_\mathrm{object}^\mathrm{shape},
    \label{eq:mask_os}
\end{align}
assuming 
$M_\mathrm{object}^\mathrm{shape} \cap M_\mathrm{person}^\mathrm{shape} = \mathbf{0}$,
where $\cap$ is an element-wise logical AND.

\subsubsection{Masking background.}

The third is ``background masking'', similar to background masking discussed in Section \ref{sec:Masking for background bias}. However, masking background (\ie, all non-human regions)
could be problematic because it removes both objects and the background from the scene.
For example, in videos of the categories ``Pretending to put [something] onto [something]'' and ``Putting [something] onto [something]'', the temporal change is caused by the presence and movement of objects, not by human actions.
To address this issue, we mask the background and objects in different colors. This eliminates bias from the object's texture while preserving the information required for category prediction.

This background masking is represented by
\begin{align}
    \begin{aligned}
    X
    =& 
    M_\mathrm{person}^\mathrm{shape} \odot V 
    + \mathbf{c}_\mathrm{bg} \!\cdot\!
        (
        \overline{M_\mathrm{person}^\mathrm{shape}}
        \cap 
        \overline{M_\mathrm{object}^\mathrm{shape}}
        )
    + \mathbf{c}_\mathrm{obj} \!\cdot\!
        (
        \overline{M_\mathrm{person}^\mathrm{shape}}
        \cap 
        M_\mathrm{object}^\mathrm{shape}
        ),
    \end{aligned}
    \label{eq:obj_bg}
\end{align}
where
$\mathbf{c}_\mathrm{obj}$ and $\mathbf{c}_\mathrm{bg}$ are colors for background and objects, respectively.

\subsection{Training procedure}

\begin{figure}[t]
    \centering
    \includegraphics[clip, width=.8\textwidth]{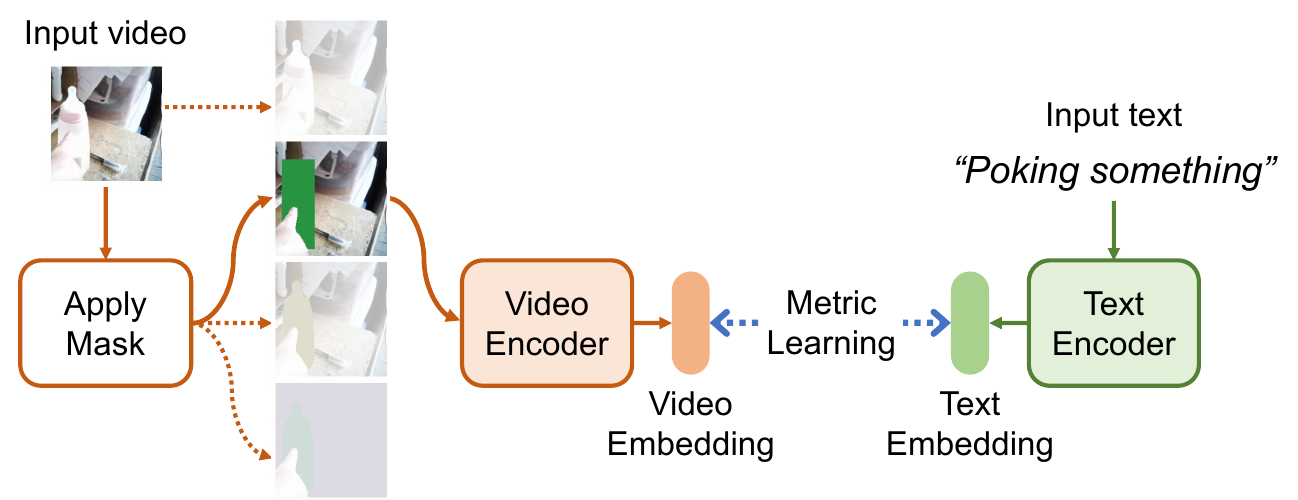}
    \caption{Training procedure with the masking approaches.}
    \label{fig:train}
\end{figure}

We implement the masking approaches through data augmentation, in a manner similar to \cite{Kimata_MMAsia2022_ObjectMix,visapp24}. The training procedure is illustrated in Figure \ref{fig:train}.
For each sample pair of video frames $V_i$ and category text $T$, we randomly apply one of the masking types with a predefined ratio (we call \emph{masking ratio}). 
For background bias, the masking ratio is represented as a number pair for choosing ``no masking'' or ``masking background''. 
For object bias, it is a quadruplet for ``no masking'', ``masking object bounding box'', ``masking object shape'', or ``masking background and object with different colors''.
Next, masked frames $X_i$ are fed into a video encoder $f_v$ to produce video embeddings $\ve_i$, while category text $T$ are passed to a text encoder $f_t$ to generate text embedding $\te$. 
Then the infoNCE \cite{Aaron_arXiv2018_infoNCE}  loss is used to learn both embeddings.

\section{Experiments}

\subsection{Settings}

\noindent\textbf{Datasets.}
For our experiments, we used three datasets:
Kinetics400 \cite{kay_arXiv2017_kinetics400} and
Mimetics \cite{Weinzaepfel_IJCV2021_Mimetics_dataset}
to assess background bias,
and
SSv2 \cite{Goyal_2017ICCV_ssv2}
to assess object bias.
In Kinetics400 \cite{kay_arXiv2017_kinetics400}, each video is sourced from YouTube and assigned to one of 400 action categories. The maximum video length is 10 seconds. The training set consists of about 240,000 videos, with each category containing between 250 and 1,000 video clips. The evaluation set includes 20,000 videos, and each category contains 50 video clips. 
Mimetics \cite{Weinzaepfel_IJCV2021_Mimetics_dataset} is an evaluation-only dataset. It consists of 713 videos across 50 categories that are a subset of Kinetics400. It includes video clips that deviate from typical scenarios, such as indoor surfing or pantomiming tennis. Therefore, these videos have less background bias and hence are challenging for models relying on the bias.
In SSv2 \cite{Goyal_2017ICCV_ssv2}, each video is recorded by crowd workers and assigned to one of the 174 template sentence categories. The average video length is 4.03 seconds and we excluded videos with shorter than 16 frames, resulting in 168,108 training videos and 24,679 validation videos.

\noindent\textbf{Model.}
We used ViFi-CLIP \cite{Rasheed_2023_ViFiCLIP} and ActionCLIP \cite{Wang_2021_ActionCLIP} as zero-shot action recognition models. ViFi-CLIP is a straightforward extension of CLIP \cite{Radford_2021_CLIP} for video, where each frame is fed to CLIP to extract features followed by aggregation with temporal average. ActionCLIP is also an extension of CLIP for video, but it performs temporal aggregation with a 6-layer transformer. We used a pretrained CLIP model of ViT-B/32 for both models and fine-tuned all parameters of both the image and text encoders.
For training, we create a video clip consisting of 8 frames uniformly sampled from each video. Unless noted otherwise, we used a batch size of 32 and AdamW \cite{Loshchilov_ICLR2019_AdamW}.
For ViFi-CLIP, the learning rate was set to 2e-6 and the number of epochs to 12.
For ActionCLIP, we used 15 epochs and the other settings of the original paper \cite{Wang_2021_ActionCLIP}.

\noindent\textbf{Masking.}
We used GroundingDINO \cite{liu_2023_grounding} and Segment Anything Model (SAM) \cite{Kirillov_2023_ICCV_SegmentAnything_SAM}, which are open-vocabulary object detection and segmentation models, to create bounding boxes and regions of objects and persons, each applied independently to every frame of a given video.
For person detection and segmentation, we used the words ``person'' and ``hand'' as prompts.
When masking objects of SSv2, we used object names as prompts, which are provided to replace placeholders [something] in the template sentence.
We applied the masking frame by frame. A frame and the prompts were fed to GroundingDINO to detect the bounding boxes of the person and objects. These boxes were then fed into SAM as prompts for person and object segmentation.
The masking ratio for background bias varied from 1:0 to 0:1 for scenarios without and with masking. Similarly, the masking ratio for object bias starts from a quadruplet of 1:0:0:0. A ratio of 1:0 (or 1:0:0:0) implies that no masking was applied during training, which is essentially the same as training the original ViFi-CLIP \cite{Rasheed_2023_ViFiCLIP} or ActionCLIP \cite{Wang_2021_ActionCLIP}, although in different training settings with our re-implementation.

\noindent\textbf{Base-to-novel zero-shot setting.}
We adopted the base-to-novel split used by \cite{Rasheed_2023_ViFiCLIP} which divides the category set into base (seen) and novel (unseen) categories. This includes three few-shot splits, with each base category containing 16 samples. The base categories are made up of the top 50\% most frequent categories in the training set, while the novel categories include the bottom 50\% least frequent categories. 
We used a learning rate of 8e-6 with 40 epochs for this setting.

\noindent\textbf{Cross-dataset fully-supervised setting.}
This setting is not zero-shot, but the datasets differ between training and validation. We trained models on the 50 categories of the Kinetics400 training set that are present in Mimetics, and then evaluated the performance on Mimetics, as in \cite{Weinzaepfel_IJCV2021_Mimetics_dataset}.

\noindent\textbf{Cross-dataset zero-shot setting.}
This setting assesses the zero-shot performance of models trained on one dataset and evaluated on another dataset containing categories not seen during training.
We trained on 350 categories of Kinetics400 which are not present in Mimetics, using a learning rate of 1e-5, and evaluated the zero-shot performance on Mimetics.

\begin{figure}[!t]
    \begin{center}
        \begin{minipage}[t]{0.45\linewidth}
            \centering
            \includegraphics[clip, width=\textwidth]{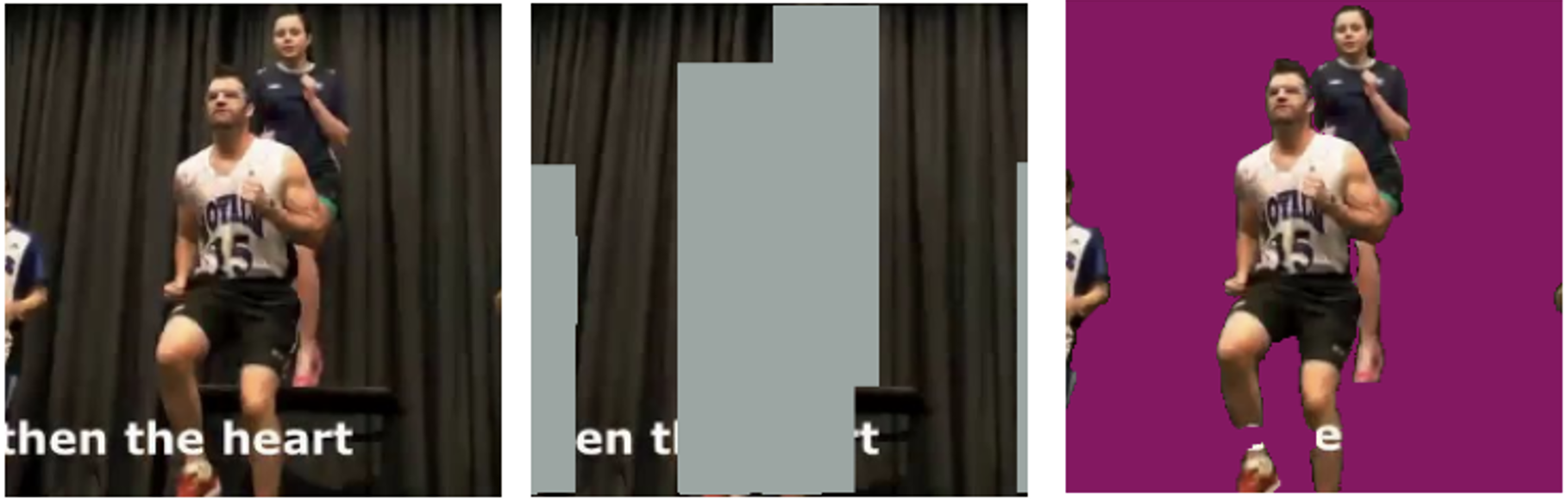}
            \subcaption{}
            \label{fig:val_mask_k400}
        \end{minipage}
        \hspace{0.05\linewidth}
        \begin{minipage}[t]{0.45\linewidth}
            \begin{center}
                \includegraphics[clip, width=\textwidth]{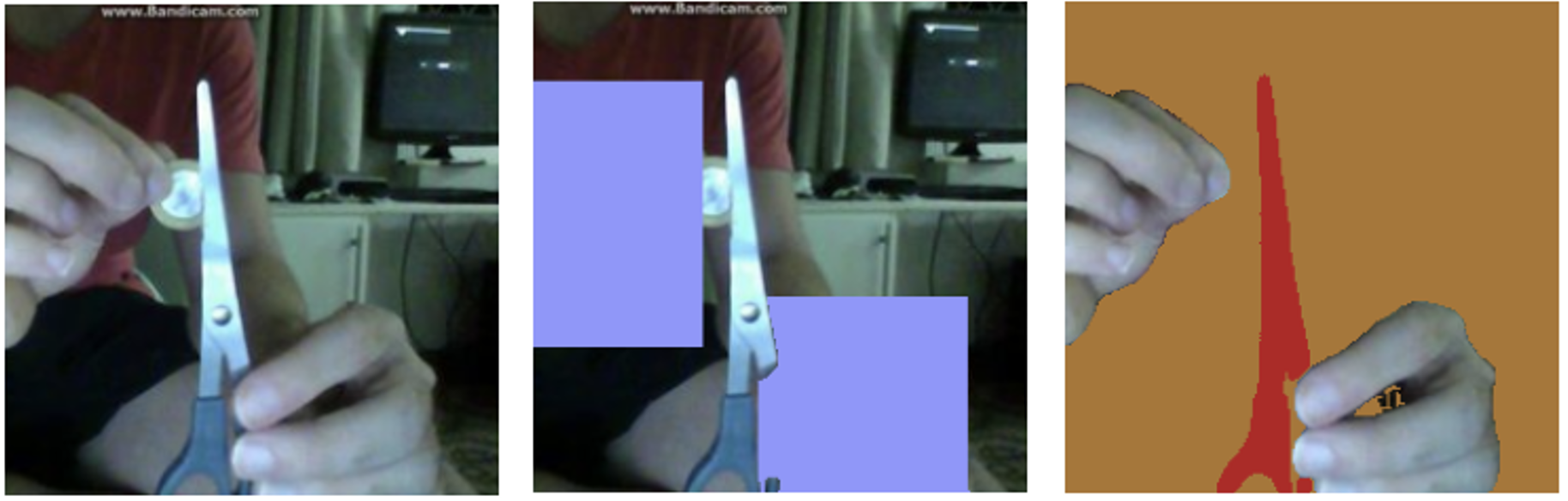}
                \subcaption{}
                \label{fig:val_mask_ssv2}
            \end{center}
        \end{minipage}
        \caption{Examples of masking for validation videos.
        Video frames from (a) Kinetics400 and (b) SSv2.
        From left to right,
        (1st) original images, masking 
        (2nd) person bounding boxes and
        (3rd) background, respectively.
        }
        \label{fig:mask_val}
    \end{center}
\end{figure}

\subsection{Evaluation metrics}

We use the top1 performance on the validation set for performance evaluation and report the best result over the training epochs (or the average across the three splits for the base-to-novel zero-shot setting). In this case, no masking is applied to the videos in the validation set; masking is used only during training.

While this is commonly used in the action recognition literature, it does not reveal any potential model bias towards the background and objects. 
To assess how much the model relies on the background,
we introduce the following two masking approaches for videos in the validation set (Figure \ref{fig:mask_val}).
\begin{itemize}

\item ``Masking background'' 
is the same with 
Eq.\eqref{eq:bg_mask} for background bias and Eq.\eqref{eq:obj_bg} for object bias.
We use the masked validation videos to compute the top1 performance, which we refer to as \emph{Person top1} or simply \emph{P-top1}. 
This term is used because, in this case, models only see the person. 

\item ``Masking person bounding box'' is intended to remove any persons from the scene.
Let $M_\mathrm{person}^\mathrm{bbox} \in \{0, 1\}^{H \times W}$ be a mask of the person bounding boxes, assigning 1 to pixels inside the bounding boxes and 0 otherwise, and the masking is represented by;
\begin{align}
    M_\mathrm{person}^\mathrm{bbox'}
    &= \overline{M_\mathrm{person}^\mathrm{bbox}}
        \cup M_\mathrm{object}^\mathrm{shape},
    \quad
    X_i
    = M_\mathrm{person}^\mathrm{bbox'} \odot V_i
    + \mathbf{c} \cdot \overline{M_\mathrm{person}^\mathrm{bbox'}}.
\end{align}
We refer to the top1 performance of this masking as \emph{Background top1} or \emph{B-top1}. This is because, similar to P-top1, models can only see the background. 
Note that a smaller B-top1 is better because a large B-top1 indicates the model depends on the background bias.

\end{itemize}

These metrics offer advantages over existing ones such as SHAcc and SBErr \cite{Action_swap} because our P-top1 and B-top1 can be calculated with lower computational cost.

\subsection{Results of masking for background bias in Kinetics}

\subsubsection{Base-to-novel zero-shot.}
Table \ref{Tab:k400b2n} shows the results of the base-to-novel zero-shot setting.
The decrease in B-top1 as the mask ratio increases indicates that the masking approach suppresses the models from learning background bias.
A decrease in top1 performance is expected because models use background bias, while masking the background during training likely prevents models from using this bias.
P-top1 was significantly lower than B-top1 without masking; however, it improved significantly with masking.
This suggests that the ability to make inferences without depending on the background information, and based solely on the information of the person, has improved.

The two models differ in the mask ratio that results in the highest P-top1, but both show that masking background of all videos does not significantly improve P-top1.
ViFi-CLIP outperformed ActionCLIP, probably because ActionCLIP trains the temporal aggregation transformer from scratch, making few-shot learning more difficult.

\begin{table}[t]
 \begin{center}

  \caption{
  Performances of base-to-novel zero-shot setting on Kinetics400 with background masking.
  The masking ratio represents a ratio between ``no masking'' (---) and ``masking background'' (mask).
  \hfill\scriptsize{${}^*$our re-implementation}
  }
  \label{Tab:k400b2n}

  \begin{tabular}{c|c@{ : }c|ccc|ccc}
    \multicolumn{1}{c}{}& \multicolumn{2}{c}{ratio} & \multicolumn{3}{c}{base (seen)} & \multicolumn{3}{c}{novel (unseen)}\\
    model
    & --- & mask & top1 & B-top1 & P-top1 & top1 & B-top1 & P-top1 \\
    \hline
    ViFi-CLIP\cite{Rasheed_2023_ViFiCLIP}${}^*$
    & 1.0 & 0.0   &  \textbf{68.87} & 50.27 & 12.08 & \textbf{51.57} & 38.60 & \phantom{0}6.91 \\ 
    & 0.5 & 0.5   &  68.24 & 49.24  & 26.08 & 50.97 & 38.44 & 13.98  \\ 
    & 0.33 & 0.67 &  67.42 & 48.40  & \textbf{26.61} & 50.58 & 37.88 & \textbf{14.08}  \\ 
    & 0.0  & 1.0  &   59.68 & 42.34 & 25.40 & 47.99 & 36.21 & 13.46 \\
    \hline
    ActionCLIP\cite{Wang_2021_ActionCLIP}${}^*$
    & 1.0  & 0.0  &  \textbf{59.11} & 42.42 & 10.00  & \textbf{30.23} & 22.67 & 5.68 \\
    & 0.5  & 0.5  &  52.92 & 37.21 & \textbf{18.40}  & 25.92 & 19.66 & \textbf{8.31} \\
    & 0.33 & 0.67 &  48.55 & 33.24 & 17.70  & 23.78 & 18.15 & 8.01 \\
    & 0.0  & 1.0  &  28.99 & 17.30 & 13.83  & 14.68 & 10.33 & 6.27 \\
  \end{tabular}

  \caption{
  Performances of cross-dataset fully-supervised setting
  where the model is
  trained on 50 classes of Kinetics400 and evaluated on the same 50 classes of Mimetics.
  }
  \label{Tab:k400_full_mim}

  \begin{tabular}{c|c@{ : }c|ccc}
    \multicolumn{1}{c}{}& \multicolumn{2}{c}{ratio} & \multicolumn{3}{c}{Mimetics} \\
    model& --- & mask & top1 & B-top1 & P-top1 \\
    \hline
    ViFi-CLIP\cite{Rasheed_2023_ViFiCLIP}${}^*$
    & 1.0 & 0.0 & 19.23 & 5.45 & 13.30\\ 
    & 0.5 & 0.5 & 22.57 & 7.81 & 21.35\\ 
    & 0.33 & 0.67 & \textbf{23.78} & 8.51 &\textbf{23.26} \\
    & 0.0 & 1.0 & 20.67 & 7.05 & 22.44\\
    \hline
    ActionCLIP\cite{Wang_2021_ActionCLIP}${}^*$
    & 1.0 & 0.0 & 20.31 & 7.47 & 16.32\\ 
    & 0.5 & 0.5 & \textbf{25.35} & 8.85 & 23.61\\ 
    & 0.33 & 0.67 & 24.31 & 9.38 & \textbf{24.13} \\
    & 0.0 & 1.0 & 21.53 & 5.90 & 23.26\\
    \hline
    S3Aug\cite{visapp24}& \multicolumn{2}{c|}{---} & 22.40 & --- & --- \\
  \end{tabular}
 \end{center}
\end{table}

\subsubsection{Cross-dataset full-supervision.}

Table \ref{Tab:k400_full_mim} shows the results of fully-supervised setting evaluated on Mimetics.
In addition to P-top1, the top1 performance improved with masking, in particular by more than 5\% for ActionCLIP. Unlike in the experiments above, top1 is improved even when masking the background of all videos. This shows that preventing the model from using background information does not negatively impact performance when the background bias of validation videos is weak.

We compared our approach with S3Aug \cite{sugiura_VISAPP2024_s3aug}, a data augmentation technique for fully-supervised action recognition that generates diverse backgrounds using segmentation and generative models, while leaving human regions intact.
Our masking approach outperformed S3Aug while requiring less computational cost, suggesting that masking is effective for backgrond bias.

\subsubsection{Cross-dataset zero-shot.}

Table \ref{Tab:k400_zs_mim} shows the results of the cross-dataset zero-shot setting evaluated on Mimetics. As expected, without masking, relying on background bias results in a higher top1 performance on Kinetics400 and poorer on Mimetics. 
In contrast, the masking approaches increased top1 on Mimetics by 1.56\% for ViFi-CLIP and 3.47\% for ActionCLIP, indicating again that the masking is effective on videos with a weak background bias.
While P-top1 improved in both datasets, B-top1 decreased in Kinetics400, indicating that the model's use of background information is being suppressed also in this setting.

\begin{table}[t]
    \begin{center}
        \caption{
          Performances of cross-dataset zero-shot setting
          where the model is
          trained on 350 classes of Kinetics400 and evaluated on both the rest 50 classes of Kinetics400 and Mimetics.
        }
        \label{Tab:k400_zs_mim}
        
  \begin{tabular}{c|c@{ : }c|ccc|ccc}
    \multicolumn{1}{c}{}& \multicolumn{2}{c}{ratio} & \multicolumn{3}{c}{Mimetics} & \multicolumn{3}{c}{Kinetics400}\\
    model& --- & mask & top1 & B-top1 & P-top1 & top1 & B-top1 & P-top1 \\
    \hline
    ViFi-CLIP\cite{Rasheed_2023_ViFiCLIP}${}^*$
    & 1.0 & 0.0   & 16.67          & 6.77 & 11.46 & \textbf{74.42} & 61.55 & 19.33 \\ 
    & 0.5 & 0.5   &17.71           & 5.38 & 15.97 & 73.60          & 61.39 & 37.99 \\ 
    & 0.33 & 0.67 & \textbf{19.27} & 6.60 & 16.50 & 72.41          & 61.23 & \textbf{42.52} \\ 
    & 0.0 & 1.0   & 18.92          & 6.25 & \textbf{16.67} & 67.64 & 55.59 & 41.69 \\ 
    \hline
    ActionCLIP\cite{Wang_2021_ActionCLIP}${}^*$
    & 1.0 & 0.0   & 16.84           & 5.90 & 13.71           & 63.73          & 52.30 & 18.38\\ 
    & 0.5 & 0.5   &  \textbf{20.31} & 8.16 & \textbf{19.44}  & \textbf{63.77} & 52.38 & 36.23\\ 
    & 0.33 & 0.67 & 20.14           & 7.64 & 17.36           & 63.32          & 50.66 & \textbf{36.60}\\ 
    & 0.0 & 1.0   &  16.15          & 5.38 & 15.62           & 52.67          & 40.87 & 34.17\\
    \end{tabular}

 \end{center}
\end{table}

\subsection{Results of masking for object bias in SSv2}

\begin{table*}[t]
 \begin{center}
  \caption{
  Performances of base-to-novel zero-shot setting on SSv2.
  The masking ratio is for 
``no masking'' (---), ``masking object bounding box'' (bbox), ``masking object shape'' (shape), and ``masking background and object with different colors'' (bg).
  }
  \label{Tab:ssv2b2n}
  \begin{tabular}{c|c@{ : }c@{ : }c@{ : }c|ccc|ccc}
    \multicolumn{1}{c}{}& \multicolumn{4}{c}{ratio} & \multicolumn{3}{c}{base (seen)} & \multicolumn{3}{c}{novel (unseen)}\\
    model
    & --- & bbox & shape & bg & top1 & B-top1 & P-top1 & top1 & B-top1 & P-top1 \\
    \hline
    ViFi-CLIP\cite{Rasheed_2023_ViFiCLIP}${}^*$
    & 1.0 & 0.0 & 0.0 & 0.0 &  12.89 & 9.36 & 5.20 & 9.62 & 7.26 & 4.40  \\ 
    & 0.5 & 0.5 & 0.0 & 0.0 &  11.78 & 8.21 & 5.77 & 8.65 & 6.11 & 4.22 \\ 
    & 0.5 & 0.0 & 0.5 & 0.0 & 12.47 & 9.03 & 7.95 & 9.45 & 6.37 & 5.55  \\ 
    & 0.5 & 0.0 & 0.0 & 0.5 & \textbf{13.93} & 9.96 & 11.01 & 9.13 & 6.50 & 6.65  \\ 
    & 0.33 & 0.0 & 0.33 & 0.33 & 13.76 & 9.45 & \textbf{11.09} & \textbf{9.64} & 6.29 & \textbf{6.89}  \\ 
    \hline
    ActionCLIP\cite{Wang_2021_ActionCLIP}${}^*$
    & 1.0 & 0.0 & 0.0 & 0.0    &  8.67 & 6.80 & 4.10  &  5.80 & 4.90 & 3.27   \\ 
    & 0.5 & 0.5 & 0.0 & 0.0    &  7.86 & 5.94 & 5.73  &  6.09 & 4.44 & 4.00 \\ 
    & 0.5 & 0.0 & 0.5 & 0.0    &  8.84 & 6.74 & 7.04  &  6.34 & 4.80 & 4.91 \\ 
    & 0.5 & 0.0 & 0.0 & 0.5    &  \textbf{11.34} & 8.35 & \textbf{10.96}  &  \textbf{8.06} & 6.09 & \textbf{7.03} \\ 
    & 0.33 & 0.0 & 0.33 & 0.33 &  10.74 & 7.73 & 10.77  &  7.50 & 5.36 & 6.61 \\ 
    
  \end{tabular}

 \end{center}
\end{table*}

Table \ref{Tab:ssv2b2n} shows the results of the base-to-novel zero-shot setting on SSv2 with masking objects and background.
The first row of each model shows the baseline performance without masking, and the next four rows show the performance with masking at different mask ratios.

\noindent\textbf{masking the object bounding box and the object shape.}
The second (0.5:0.5:0:0) and third (0.5:0:0.5:0) rows show performances with masking the object bounding box and the object shape, respectively. In both cases, background information is available during the training.
B-top1, which is the performance when the background (and also the objects) are available during evaluation, decreases in all cases. This suggests the following.
First, the difference in B-top1 with and without masking is marginal ($\pm$ 1\%), even when background information is available during both training and evaluation. This means that there is no clue in the background (i.e., weak background bias).
Second, masking objects during training leads to poorer performance, and the object appearances available during evaluation do not help. Therefore, object appearance may serve as an important clue (i.e., a strong object bias).
Third, masking the object bounding box is worse than masking object shapes. Hence, the shape of the object is a cue for prediction, and hiding it may be useful for object bias.

\noindent\textbf{masking the object and background in different colors.}
The fourth (0.5:0:0:0.5) row shows the performances with masking the object and background in different colors.
Since the background is masked but the object shape is available, the performance is expected to remain the same as when only the object shape is masked. 
However, the performance is higher than expected. This might be due to the masking approach. By removing unnecessary background information, models can focus directly on the shape of the objects.

\noindent\textbf{a combination of masking.}
The fifth row (0.33:0:0.33:0.33) shows the performance of a combination of two masking approaches,
which is a similar or higher performance to the fourth row. During training, we use both masking object shapes (but the background is available) and masking background and objects (but object shape is available). This combination of masking approaches might provide useful information to the models and lead to better performance.

\noindent\textbf{model architecture difference.}
The fourth row shows the best top-1 performance for the seen categories in most cases. However, the best performance for unseen categories is achieved at different mask ratios for different models. In the fourth row, ActionCLIP performance improved by 2\%, while ViFi-CLIP performance decreased by 0.5\% from baseline. Therefore, mask ratios or masking approaches need to be tuned differently for different model architectures.

\section{Conclusion}

In this study, we addressed the issue of static bias in zero-shot action recognition by investigating the impact of background and object biases on model performance. We tackled this issue by masking backgrounds and objects and evaluated their effectiveness on Kinetics400, Mimetics, and SSv2.
Our experimental results showed that zero-shot models depend too much on background bias. Masking the background during training prevents the model from focusing excessively on this bias, as evidenced by the improvement of P-top1.
Models also rely on object bias. Masking objects effectively reduced the model's reliance on this bias, demonstrated by the improved performance when both backgrounds and objects were masked with different colors.
Future work may include devising a model based on the results that incorporates a structure to mask the background or objects, thereby focusing more on human actions.